\documentclass[10pt,twocolumn,letterpaper]{article}

\usepackage{cvpr}

\usepackage{multirow}

\definecolor{cvprblue}{rgb}{0.21,0.49,0.74}
\usepackage[pagebackref,breaklinks,colorlinks,allcolors=cvprblue]{hyperref}

\def\confName{}
\def\confYear{}

\title{Beyond Motion Pattern: An Empirical Study of Physical Forces for Human Motion Understanding}

\author{
Anh Dao\textsuperscript{1}\thanks{Equal contribution} \qquad
Manh Tran\textsuperscript{1}\footnotemark[1] \qquad
Yufei Zhang\textsuperscript{2} \qquad
Xiaoming Liu\textsuperscript{1} \qquad
Zijun Cui\textsuperscript{1} \\
\\
\textsuperscript{1}Michigan State University \quad \textsuperscript{2}Independent Researcher\\
{\tt\small \{anhdao, tranman1, liuxm, cuizijun\}@msu.edu, yufeizhang96@outlook.com}
}

\usepackage{tcolorbox}
\tcbuselibrary{skins} 

\usepackage[svgnames,x11names,dvipsnames]{xcolor}
\tcbset{before skip=6pt, after skip=6pt}

\definecolor{cvprBlue}{HTML}{0072B2}   
\definecolor{cvprOrange}{HTML}{E69F00} 
\definecolor{cvprGreen}{HTML}{009E73}  

\newif\ifcvprbw
\ifcvprbw
  \definecolor{cvprBlue}{gray}{0.85}
  \definecolor{cvprOrange}{gray}{0.80}
  \definecolor{cvprGreen}{gray}{0.82}
\fi

\usepackage{xcolor}
\tcbuselibrary{skins,breakable} 

\definecolor{cvprBlue}{HTML}{0072B2}
\definecolor{cvprOrange}{HTML}{E69F00}
\definecolor{cvprGreen}{HTML}{009E73}

\newif\ifcvprbw
\ifcvprbw
  \definecolor{cvprBlue}{gray}{0.85}
  \definecolor{cvprOrange}{gray}{0.80}
  \definecolor{cvprGreen}{gray}{0.82}
\fi

\newtcolorbox{keyfindingbox}[2][]{%
  enhanced, breakable, sharp corners,
  colback=cvprBlue!6, colframe=cvprBlue!55!black,
  boxed title style={colback=cvprBlue!12, boxrule=0pt, sharp corners},
  borderline west={2.2pt}{0pt}{cvprBlue!85!black},
  left=1.2mm, right=1.2mm, top=1mm, bottom=1mm,
  before skip=6pt, after skip=6pt,
  title={\textbf{#2}}, fonttitle=\bfseries,
  #1
}

\begin{document}
\maketitle
\begin{abstract}
Human motion understanding has advanced rapidly through vision-based progress in recognition, tracking, and captioning. However, most existing methods overlook physical cues such as joint actuation forces that are fundamental in biomechanics. This gap motivates our study: if and when do physically inferred forces enhance motion understanding? By incorporating forces into established motion understanding pipelines, we systematically evaluate their impact across baseline models on 3 major tasks: gait recognition, action recognition, and fine-grained video captioning. Across 8 benchmarks, incorporating forces yields consistent performance gains; for example, on CASIA-B, Rank-1 gait recognition accuracy improved from 89.52\% to 90.39\% (+0.87), with larger gain observed under challenging conditions: +2.7\% when wearing a coat and +3.0\% at the side view. On Gait3D, performance also increases from 46.0\% to 47.3\% (+1.3). In action recognition, CTR-GCN achieved +2.00\% on Penn Action, while high-exertion classes like punching/slapping improved by +6.96\%. Even in video captioning, Qwen2.5-VL’s ROUGE-L score rose from 0.310 to 0.339 (+0.029), indicating that physics-inferred forces enhance temporal grounding and semantic richness. These results demonstrate that force cues can substantially complement visual and kinematic features under dynamic, occluded, or appearance-varying conditions.

\end{abstract}

\section{Introduction}
\label{sec:intro}

Understanding human motion is a long standing challenge in computer vision, covering a wide range of applications from biometric \cite{azad2024activity}, action recognition \cite{kong2022human}, video captioning \cite{xu2024finesports}, and human-AI interaction \cite{teo2019mixed, chakraborty2018review}. Despite substantial progress fueled by deep learning, existing approaches to human motion understanding remain largely appearance or kinematics-based, each with inherent limitations. 

Appearance-based methods rely on visual cues such as silhouettes or images \cite{chao2019gaitset, fan2020gaitpart, fan2023opengait, simonyan2014two, carreira2017quo, fan2018end, bertasius2021space, tong2022videomae}, which encode coarse body outlines but suffer when appearance varies due to clothing or camera viewpoints. On the other hand, kinematics-based models capture geometric motion patterns through 2D or 3D joint estimations \cite{chen2021channel, teepe2022towards, teepe2021gaitgraph}. However, they are limited by frequently observed physically implausible estimates, such as motion jitter, which lead to distorted spatiotemporal structures and degraded performance \cite{liu2020disentangling, plizzari2021skeleton}. Moreover, they remain agnostic to the underlying causal dynamic information, such as forces that account for body movements. These weaknesses show a fundamental limitation in current paradigms: they describe how motion looks or moves, but not \textit{why} it occurs, failing to capture the underlying physical dynamics that drive human behavior.

\begin{figure}[t!]
    \centering
    \includegraphics[width=1.0\linewidth]{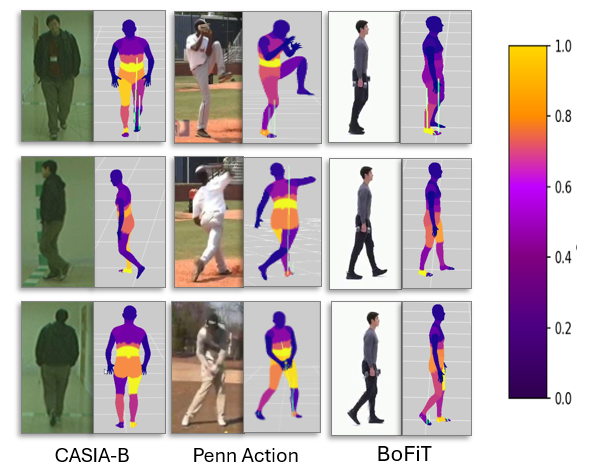}
    \caption{Visualization of RGB frames and their corresponding motion forces from multiple datasets used in gait recognition, action recognition, and video captioning tasks.  The magnitude of joint actuations is encoded by color at each body part, with lighter colors indicating higher magnitudes.}
    \label{fig:fig1}
\end{figure}

Human movement is fundamentally produced by physical forces, which reveal an individual’s stability, efficiency, and motor control. As illustrated in Figure \ref{fig:fig1}, these forces reveal motion characteristics beyond what appearance or pose alone convey. The first column presents three viewpoints of the same walking sequence, where weight transfer patterns remain consistent across viewpoints, benefiting gait recognition by providing view-invariant locomotion cues. The second column shows concentrated contact forces and joint actuations around supporting limbs and the torso, depicting how momentum shifts as the body leans, pushes, or interacts with the ground. The final examples capture subtle variations in exertion and coordination across the body, exposing fine-grained cues related to motion realism.

Leveraging physical forces is rarely explored in human motion understanding, as capturing and integrating hidden physical signals remains non-trivial. Recently, several studies have shown promising progress by grounding motion in physics, including physically consistent pose estimation \cite{rempe2020contact, li2019estimating} and dynamics-driven character control \cite{gartner2022differentiable, levine2012physically}. In light of the recent success of physics-driven advances and human dynamics estimation~\cite{zhang2024physpt}, we study the effectiveness of physical forces across a range of human motion understanding tasks, including gait recognition, action recognition, and fine-grained video captioning.

We present an systematic investigation into the role of force as a new modality, including its representation learning and integration with various leading models for human motion understanding. We effectively encode joint actuation forces into compact and discriminative features that complement conventional visual cues. Extensive experiments on multiple benchmarks demonstrate the advantages of our approach, showing consistent improvement under different challenging covariates such as viewpoint variation, clothing changes, and occlusion. We hope our method, by bridging the gap between appearance or kinematics based approaches and physics informed modeling, establishes a strong baseline for future research. In summary, our main contributions include: 
\begin{itemize}
    \item An introduction of physical motion forces as a novel modality and their integration into strong baselines for human motion analysis.
    \item A comprehensive evaluation of this modality on eight benchmarks across three distinct tasks: gait recognition, action recognition, and fine-grained video captioning.
    \item Empirical insights and deep analysis that answer \textit{when} and \textit{why} dynamics help (e.g., under appearance variation, for high-exertion actions) and when they do not (\textit{e.g.}, due to modality redundancy).
\end{itemize}

\section{Related Work}
\label{sec:related_work}

We categorize related work on human motion understanding based on their motion representations into three groups: appearance-based, kinematics-based, and physics-informed approaches. Our discussion covers three key tasks: gait recognition, action recognition, and video captioning.

\noindent\textbf{Appearance-based Approaches} rely on extracting visual features from input images. In gait recognition, current methods strive to leverage silhouettes and their combination with RGB image features extracted by deep learning models~\cite{chao2019gaitset, fan2020gaitpart, fan2023opengait, zhang2020learning, ye2024biggait, ye2025biggergait, lin2022gaitgl, ma2023dynamic,fan2023learning, wang2024qagait}. In action recognition, appearance-based pipelines include early handcrafted descriptors~\cite{scovanner20073, al2012spatio, klaser2008spatio} such as SIFT~\cite{laptev2005space}, followed by deep learning methods like 3D CNNs~\cite{tran2015learning, simonyan2014two} and transformer-based models~\cite{bertasius2021space, tong2022videomae, arnab2021vivit, yang2022recurring, yamane2024mvaformer}. Beyond recognition, appearance-based cues have also been extended to motion captioning, where models generate natural language descriptions of human movement. Early studies~\cite{takano2015statistical, fan2018end, yamada2018paired} employed statistical models to map motion sequences to text, while recent vision-language frameworks~\cite{zhao2025exploring, lin2024video, li2023videochat} achieve strong performance by training on large-scale video–text pair. Nonetheless, appearance-based methods remain visually centered, making them highly sensitive to covariates such as clothing changes, carrying conditions, occlusions, and viewpoint shifts.

\noindent\textbf{Kinematics-based Approaches} focus on motion geometry such as joint positions and velocities. For gait recognition, existing methods have employed skeleton pose information with dedicated graph neural networks (GCNs) to model spatiotemporal dependencies between joints~\cite{teepe2021gaitgraph, teepe2022towards, zhang2023spatial, peng2024learning, fu2023gpgait}. Parametric body models~\cite{loper2023smpl} have also been utilized to represent body geometry, combined with heatmaps of joint coordinates for more discriminative gait modeling~\cite{fan2024skeletongait, zheng2022gait}. In action recognition, skeleton-based approaches dominate large-scale benchmarks such as NTU-RGB+D~\cite{shahroudy2016ntu}, where GCNs are particularly effective \cite{zhou2024blockgcn, chi2022infogcn, liu2020disentangling, chen2021multi, lee2023hierarchically, song2022constructing, chen2021channel}, such as CTR-GCN\cite{chen2021channel} refines skeleton topology channel-wise to better capture dynamics. For video captioning, kinematics cues have been incorporated into vision-language frameworks to improve motion-grounded reasoning~\cite{chen2024motionllm}. However, kinematics-based methods do not model the underlying forces that drive motion and therefore cannot explain why it occurs, which limits their performance gains.

\noindent\textbf{Physics-informed Approaches} explicitly incorporate physical laws or model latent forces, torques, and contact dynamics. These methods have demonstrated promising results, such as enhancing physical plausibility in character control systems~\cite{gartner2022differentiable, levine2012physically, shimada2020physcap, yi2022physical}, advancing 3D human motion estimation~\cite{rempe2020contact, li2019estimating, xie2021physics} and forecasting~\cite{Zhang_2024_WACV}, and delivering strong monocular human dynamics estimation~\cite{zhang2024physpt}. We extend this line of work in pure motion estimation by focusing on leveraging physical forces across three important tasks: gait recognition, action recognition, and motion captioning. Furthermore, Guo \textit{et al.}~\cite{guo2023physics} introduced physics-augmented neural networks to implicitly capture physical priors. In contrast, we explicitly introduce physical forces as an additional modality, showing that this explicit formulation can complement SoTA models for human motion understanding.

\section{Methodology}
\label{sec:methodology}

\begin{figure*}[t]
    \centering
    \includegraphics[width=0.8\textwidth]{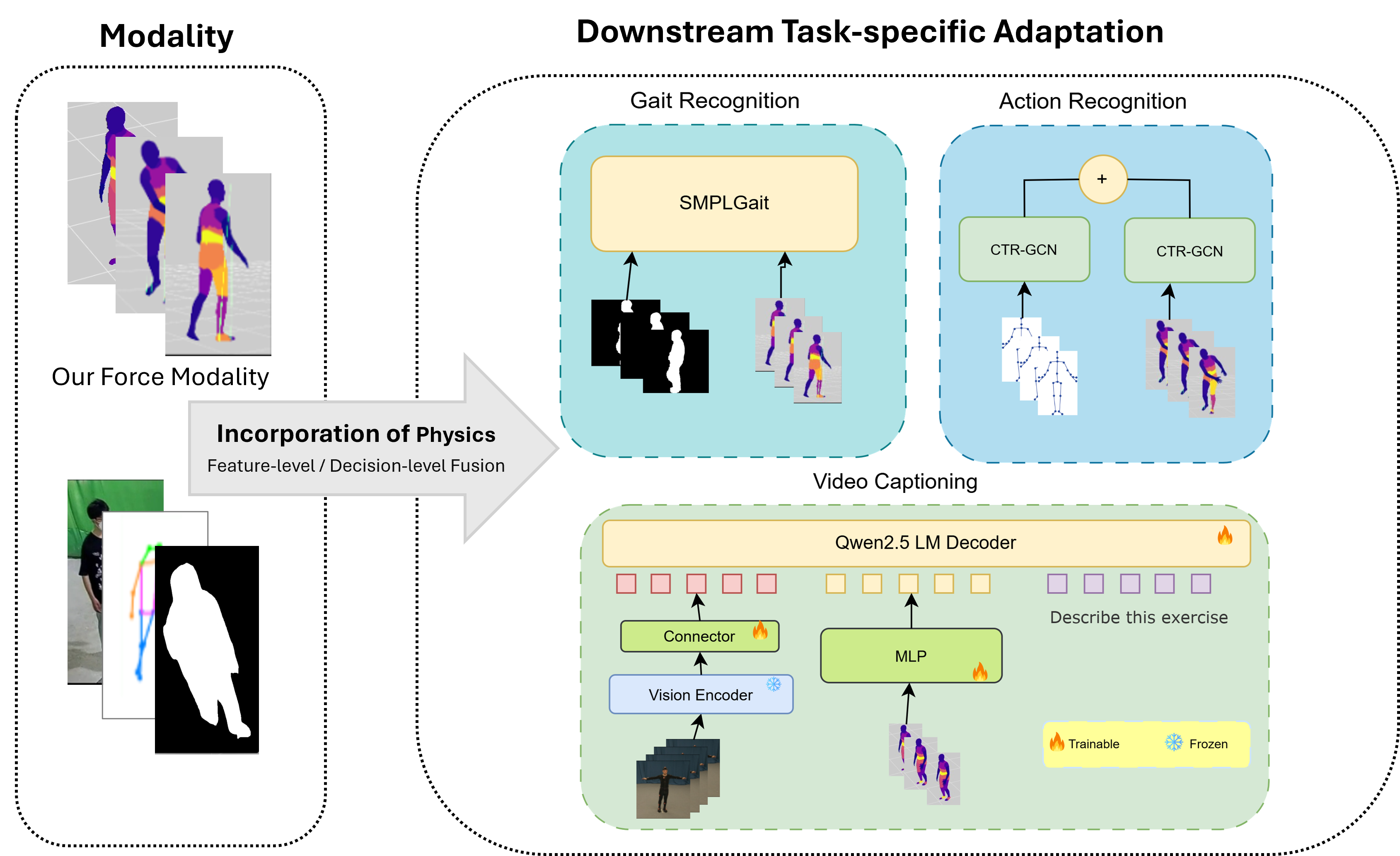}
    \caption{Overview of our framework. Force estimated from video via physics-informed model are fused with conventional visual modalities (RGB, pose, silhouette) through feature-level or decision-level fusion, and then used for task-specific adaptation.}
    \label{fig:pipeline3}
\end{figure*}

We present in Figure~\ref{fig:pipeline3} an overview of our framework, which investigates the integration of force as a new modality into established models for human motion analysis. In the following, we begin by defining the standard data modalities. Then, we introduce in Section~\ref{sec:forcemodality} our proposed force representation derived from physics principles. We present in Section~\ref{sec:fusionstategy} a generalized taxonomy of multi-modal fusion strategies, which is later instantiated for three downstream tasks.

For a video sequence of $T$ frames, the widely used modalities in existing models include:
\begin{itemize}
 \item \textbf{Appearance:} The raw video frames, $\mathcal{V} = \{V_t \in \mathbb{R}^{H \times W \times 3}\}_{t=1}^T$, where $H \times W$ are the image dimensions. In gait recognition, \textbf{silhouettes} are commonly used, extracted from images as binary foreground masks, $\mathcal{S} = {S_t \in [{0,1}]^{H \times W}}_{t=1}^T$. Appearance-based features capture 2D visual information but are sensitive to covariates such as clothing and viewpoint. 
 \item \textbf{Kinematics:} 3D joint positions or pose parameters (e.g., from SMPL~\cite{loper2023smpl}), $\mathcal{K} = \{K_t \in \mathbb{R}^{D}\}_{t=1}^T$, where $D$ represents the dimensionality of the positions or parameters. Kinematics describe the geometric patterns of motion but do not capture the underlying causal dynamics.
\end{itemize}

\subsection{Force as a New Modality}
\label{sec:forcemodality}
While appearance and kinematics describe \textit{how} a person moves, they do not explicitly model the underlying dynamics that explain \textit{why} they move. To bridge this gap, we introduce force as a new modality.

\subsubsection{Joint Torques.}
Human dynamics are widely studied and described by the Euler–Lagrange equations:
\begin{equation}
    \mathbf{M}(\mathbf{q})\ddot{\mathbf{q}} + \mathbf{C}(\mathbf{q}, \dot{\mathbf{q}}) + \mathbf{g}(\mathbf{q}) = \mathbf{J}_C^\top \boldsymbol{\lambda} + \boldsymbol{\tau},
\end{equation}
where $\mathbf{q}$ represents the generalized coordinates (e.g., joint angles and global translation) that fully specify a motion state, $\mathbf{M}(\mathbf{q})$ is the generalized mass matrix, $\mathbf{C}(\mathbf{q}, \dot{\mathbf{q}})$ accounts for Coriolis and centrifugal forces, $\mathbf{g}(\mathbf{q})$ is the gravitational force, $\mathbf{J}_C^\top \boldsymbol{\lambda}$ captures external contact forces, and $\boldsymbol{\tau}$ are the internal joint actuation torques.

In this work, we focus on the joint torques $\boldsymbol{\tau}$, as they directly reflect the subject's biomechanical effort and neuromuscular control. As discussed in Section~\ref{sec:intro} with the illustration Figure~\ref{fig:fig1}, the magnitude and direction of these torques effectively reveal where and how strongly the body exerts effort. Therefore, we introduce the torque modality as a significant and informative representation that captures the essential dynamics of human motion:
\begin{itemize}
    \item \textbf{Dynamics (Force):} Joint torques $\boldsymbol{\tau} = \{ \boldsymbol{\tau}_t \in \mathbb{R}^{J \times 3} \}_{t=1}^T$, where $J$ represents the number of body joints.
\end{itemize}
During implementation, we estimate $\boldsymbol{\tau}$ from monocular video sequences using PhysPT~\cite{zhang2024physpt}, without requiring ground-truth force data.

\subsubsection{Force Representation}
The dynamics or force discussed above provides a complementary modality to appearance and kinematics, enhancing robustness across various tasks. We introduce a modality-specific \textbf{Force Network (FN)} that processes the torque signals $\boldsymbol{\tau} = \{\boldsymbol{\tau}_t \in \mathbb{R}^{J\times 3}\}_{t=1}^T$ to extract a compact feature embedding:
\begin{equation}
\phi_{\text{force}} = \mathrm{FN}(\boldsymbol{\tau}; \theta_f)
\label{eq:force_network}
\end{equation}
where the network $\mathrm{FN}(\cdot; \theta_f)$, with learnable parameters $\theta_f$, can be implemented as a multilayer perceptron (MLP) or graph convolutional layers.

\subsection{Multi-Modal Fusion Strategies}
\label{sec:fusionstategy}
To effectively integrate the force representation $\phi_{\text{force}}$ with traditional modalities, we explore different fusion strategies, including feature-level and decision-level fusion.

\subsubsection{Feature-Level Fusion}
This is the most common and flexible approach, where features from different modality-specific branches are combined at one or more intermediate layers. Given feature representations from two modalities, $\phi_A$ and $\phi_B$, extracted by their respective networks, a fusion operator $\mathcal{F}$ generates a unified representation:
\begin{equation}
 \phi_{\text{fused}} = \mathcal{F}(\phi_A, \phi_B)    
\end{equation}
The operator $\mathcal{F}$ can be a simple parameter-free function like concatenation, addition, or element-wise multiplication. Alternatively, it can be a learnable module, such as a cross-attention mechanism, a gating network, or a compact bilinear pooling layer, which can model more complex inter-modal relationships.

In particular, a specialized form of intermediate fusion is employed in multi-modal large vision language models (VLMs). Here, embeddings from different modalities (e.g., vision $\phi_{\text{vis}}$, text $\phi_{\text{text}}$, force $\phi_{\text{force}}$) are projected into a common feature space and then concatenated to form a single input sequence for the LLM's decoder.
\begin{equation}
\label{eq:fusionVLM}
\phi_{\text{fused-VLM}} = [\text{Proj}(\phi_{\text{vis}}); \text{Proj}(\phi_{\text{text}}); \text{Proj}(\phi_{\text{force}})]
\end{equation}
This allows the model's self-attention mechanism to fluidly arbitrate between and combine information from all modalities when generating a sequential output like a text caption.

\subsubsection{Decision-Level Fusion}
\label{sec:decisionfusion}
This strategy involves training separate, independent models for each modality all the way to their final outputs (e.g., classification scores). The final prediction $P_{\text{fused}}$ is produced by combining these individual outputs:
\begin{equation}
\label{eq:fusionweight}
    P_{\text{fused}} = w_A P_A + w_B P_B 
\end{equation}
where $P_A$ and $P_B$ are the output from each model, and $w_A, w_B$ are scalar weights (e.g., $0.5$ for mean fusion). This approach is simple to implement and allows each stream to become a specialized expert on its modality.


\section{Force Dynamics across Tasks}

In this section, we detail how the force representation (introduced in Section~\ref{sec:forcemodality}) and the general fusion strategies are instantiated within state-of-the-art (SOTA) architectures for gait recognition, action recognition, and video captioning.

\subsection{Gait Recognition}
\label{sec:gait_task}

We integrate force into the SMPLGait~\cite{zheng2022gait} architecture, a strong baseline combining appearance and kinematic cues.

\noindent\textbf{Architecture.}
The original SMPLGait employs a Silhouette Learning Network (SLN) for appearance features and a 3D Spatial-Transformation Network (3D-STN) for SMPL-based kinematics. We introduce a third branch, the Force Network (FN), implemented as an MLP that encodes the estimated joint-actuation forces into an embedding $\phi_{\text{force}}$.
\paragraph{Fusion Strategy.}
To maintain compatibility with SMPLGait’s mid-level fusion design, we adopt \textbf{Feature-Level Fusion} where $\phi_{\text{force}}$ is projected and spatially aligned with both the silhouette ($\phi_{\text{sil}}$) and kinematic ($\phi_{\text{kin}}$) feature maps. The fused representation is learned through a spatial transformation module (BMM) that adaptively reweights local appearance features based on the underlying dynamics. The fused features are then aggregated through global pooling and passed to the classifier for final identity prediction.

\subsection{Action Recognition}
We consider skeleton-based action recognition and adopt a leading graph convolutional networks, CTR-GCN~\cite{chen2021channel}.

\noindent\textbf{Architecture.} 
To align with the CTR-GCN~\cite{chen2021channel} architecture, we introduce an identical model that operates on the skeleton graph but in the force domain to process joint torques $\boldsymbol{\tau}$. 
This stream is designed to capture human dynamics.

\noindent\textbf{Fusion Strategy.}
Both kinematic and dynamic information contribute significantly to action recognition, even when used independently. To isolate and quantify the predictive power of the force modality, we adopt \textbf{Decision-Level Fusion} for this task, setting the fusion weights to $w_A = w_B = 0.5$ in Equation~\ref{eq:fusionweight}. This strategy treats the kinematics-based and force-based branches as independent "experts," each trained to specialize in its own modality, resulting in a strong, simple, and interpretable model.  

\subsection{Fine-Grained Video Captioning}
We build upon powerful Large Vision-Language Models (LVLMs), such as Qwen2.5-VL~\cite{bai2025qwen2}, to assess the utility of force in generative language tasks.

\noindent\textbf{Architecture.}
Qwen2.5-VL features a vision encoder and an LLM decoder, designed to process interleaved visual and textual tokens. On top of the estimated dynamics $\boldsymbol{\tau}$, we apply a MLP projector to obtain a compact force embedding $\phi_{\text{force}}$, which is then fed into the LLM for joint reasoning.

\noindent\textbf{Fusion Strategy.}
The LLM decoder is designed to operate on a unified sequence of token embeddings. Therefore, we employ \textbf{Feature-Level Fusion} with the strategy specifically tailored for VLMs, as introduced in Section~\ref{sec:fusionstategy} and Equation~\ref{eq:fusionVLM}. Specifically, we project the force embedding $\phi_{\text{force}}$ into the same latent space as the visual $\phi_{\text{vis}}$ and textual $\phi_{\text{text}}$ embeddings, treating force as a new type of input token. This direct access to dynamic cues enables the model to generate more physically plausible and descriptive captions, capturing subtle aspects of effort, balance, and execution beyond visual or kinematic information.
\section{Experimental Setup}
In this section, we present the evaluation protocols, datasets, and metrics for the three tasks. Additional implementation details are provided in Appendix~\ref{app:impl}.

\subsection{Gait Recognition}

\noindent\textbf{Datasets.}
We evaluate on three benchmarks. Gait3D~\cite{zheng2022gait} is a large-scale in-the-wild dataset with 4{,}000 subjects and 25{,}000+ sequences captured by 39 cameras; it provides SMPL models recovered from video frames. We follow its official train/test protocol. CASIA-B~\cite{yu2006framework} contains 124 subjects under three covariates—normal walking (NM), walking with a bag (BG), and wearing a coat (CL)—and we adopt the common split (first 74 subjects for training, remaining 50 for testing). CCGR-Mini~\cite{zou2024cross} is a compact cross-covariate benchmark with 970 subjects and 47{,}884 sequences, retaining 53 covariates (one view per covariate) to speed up experimentation.

\noindent\textbf{Evaluation Metrics.}
Following standard practice, we report Rank-1/5/10 identification accuracy and retrieval metrics (mAP and mINP). On CASIA-B, we further analyze performance under NM/BG/CL. 

\subsection{Action Recognition}
Our action recognition experiments follow the standard bone-based input setting. Specifically, we adopt the bone modality as the kinematic representation and introduce force dynamics as an additional input stream on top of it. We report results using the bone-only and bone-force configurations to isolate the contribution of the force modality.

\noindent\textbf{Datasets.}
We use four benchmarks. NTU-RGB+D 60~\cite{shahroudy2016ntu} contains 56{,}880 clips across 60 classes with Cross-Subject (XSub) and Cross-View (XView) protocols. NTU-RGB+D 120~\cite{liu2019ntu} extends this to 114{,}480 clips and 120 classes with Cross-Subject (XSub) and Cross-Setup (XSet) protocols, where training and testing are conducted on disjoint camera setups to evaluate generalization to unseen environments. NW-UCLA~\cite{wang2014cross} features 1{,}494 videos of 10 actions captured by three Kinect cameras. Penn Action~\cite{zhang2013actemes} includes 2{,}326 videos of 15 actions with 2D joint annotations.

\noindent\textbf{Evaluation Metrics.}
For NTU-60, we follow the official XSub and XView protocols, and for NTU-120, the XSub and XSet protocols. In all cases, we report top-1 classification accuracy, computed as the percentage of correctly predicted action classes on the test set for each protocol. For NW-UCLA and Penn Action, we also report top-1 accuracy following prior work.

\subsection{Fine-Grained Video Captioning}
We adopt Qwen2.5-VL-3B~\cite{bai2025qwen2} for video captioning, consisting of a frozen vision encoder, a trainable connector, and an LLM decoder. Visual features from the encoder are projected through the connector into the language space. The force dynamics $\boldsymbol{\tau}$ are encoded via a trainable MLP projector into an embedding $\phi_{\text{force}}$, which is fused with visual features before decoding. Only the connector, MLP, and decoder are finetuned, while the vision encoder remains frozen.

\noindent\textbf{Dataset.}
We use BoFiT~\cite{zhao2025exploring}, a fine-grained fitness-training captioning dataset with 2{,}360 videos and detailed step-level descriptions. As the original paper does not prescribe an official split, we adopt a stratified split by exercise type and difficulty (2{,}000 train / 360 test); we release the split strategy in the Appendix~\ref{app:data}.

\noindent\textbf{Evaluation Metrics.}
We evaluate using ROUGE-L~\cite{lin2004rouge} and BERTScore~\cite{zhang2019bertscore}. ROUGE-L measures overlap and recall between generated and reference captions, while BERTScore assesses semantic similarity by computing contextual embeddings from a pretrained BERT model. 

\section{Results \& Analyses}
\label{sec:results}

We present a systematic evaluation of our proposed force modality across gait recognition, action recognition, and video captioning. Our analysis first establishes the overall effectiveness of force dynamics (Section~\ref{sec:overall_perf}), then examines its robustness to domain shifts (Section~\ref{sec:robustness}) and design choices through ablation studies (Section~\ref{sec:ablation}).

\subsection{Overall Performance Analysis}
\label{sec:overall_perf}

\begin{table*}[t]
\centering
\caption{\textbf{Overall Performance across Three Tasks.} Integrating force dynamics (`+ Force`) consistently improves performance over strong baselines across all three tasks. The baselines are Silhouette-Only (gait), Kinematics-Only (action), and Vision-Only (captioning). Best results are in bold, and $\Delta$ indicates absolute improvement.}
\label{tab:overall_performance}
\small
\begin{tabular}{@{}lllccc@{}}
\toprule
\textbf{Task} & \textbf{Dataset} & \textbf{Metric} & \textbf{Baseline} & \textbf{+ Force (Ours)} & \textbf{$\Delta$} \\
\midrule
\multirow{5}{*}{\textbf{Gait Recognition}} 
& \multirow{2}{*}{Gait3D} & Rank-1 (\%) & 46.0 & \textbf{47.3} & +1.3 \\
& & mAP (\%) & 38.02 & \textbf{39.11} & +1.09 \\
\cmidrule(l){2-6}
& \multirow{2}{*}{CCGR-mini} & Rank-1 (\%) & 19.3 & \textbf{20.6} & +1.3 \\
& & mAP (\%) & 19.3 & \textbf{20.7} & +1.4 \\
\cmidrule(l){2-6}
& CASIA-B & Rank-1 (Avg.) (\%) & 89.52 & \textbf{90.39} & +0.87 \\
\midrule
\multirow{6}{*}{\textbf{Action Recognition}}
& \multirow{2}{*}{NTU-60} & X-Sub Acc. (\%) & 89.53 & \textbf{89.96} & +0.43 \\
& & X-View Acc. (\%) & 94.36 & \textbf{94.90} & +0.54 \\
\cmidrule(l){2-6}
& \multirow{2}{*}{NTU-120} & X-Sub Acc. (\%) & 85.55 & \textbf{85.86} & +0.31 \\
& & X-Set Acc. (\%) & \textbf{85.12} & 84.31 & -0.81 \\
\cmidrule(l){2-6}
& NW-UCLA & Accuracy (\%) & 93.00 & \textbf{93.97} & +0.97 \\
\cmidrule(l){2-6}
& Penn Action & Accuracy (\%) & 96.00 & \textbf{98.00} & +2.00 \\
\midrule
\multirow{2}{*}{\textbf{Video Captioning}}
& \multirow{2}{*}{BoFiT} & ROUGE-L & 0.310 & \textbf{0.339} & +0.029 \\
& & BERTScore & 0.8921 & \textbf{0.8958} & +0.0037 \\
\bottomrule
\end{tabular}
\end{table*}

We observe consistent gains from incorporating force dynamics across all three tasks. Table~\ref{tab:overall_performance} summarizes results on 8 benchmarks for a unified comparison of the force modality's contribution. For gait recognition, force improves Rank-1 accuracy by 0.87--1.4\% across all datasets, demonstrating its value in person re-identification. For action recognition, it enhances performance on 5 of 6 benchmarks, including NTU-60 X-Sub (+0.43\%), NTU-60 X-View (+0.54\%), NTU-120 X-Sub (+0.31\%), Penn Action (+2.00\%), and NW-UCLA (+0.97\%), while slightly decreasing on NTU-120 X-Set (–0.81\%), as further analyzed in Section~\ref{sec:view_vs_subject}. For video captioning, force improves both ROUGE-L and BERTScore, indicating richer, semantically grounded descriptions through the addition of physics-based context.


\subsection{Robustness Analysis}
\label{sec:robustness}
Having established the general utility of force, we now probe its behavior under challenging conditions to understand \textit{when} and \textit{why} it is most beneficial.

\subsubsection{Robustness to Appearance and Viewpoint Changes}
\label{sec:appearance_invariance}

\begin{table}[t!]
\centering
\caption{\textbf{Gait Recognition Robustness Evaluation on CASIA-B.} Force provides the largest benefits under challenging conditions, where appearance is altered (CL) and the viewpoint is non-frontal (e.g., $90^{\circ}$). Rank-1 accuracy (\%) is reported using BMM fusion. The best result for each condition is highlighted in \textbf{bold}.}
\label{tab:casia_analysis_detailed}
\small
\begin{tabular}{@{}llccc@{}}
\toprule
\textbf{Condition} & \textbf{Probe View} & \textbf{Sil-Only} & \textbf{+ Force} & \textbf{$\Delta$} \\
\midrule
\multirow{4}{*}{Normal (NM)} 
& $0^{\circ}$--$180^{\circ}$ (All) & 96.12 & \textbf{96.95} & +0.83 \\
\cmidrule(l){2-5}
& $0^{\circ}$ (Frontal) & 89.50 & \textbf{92.80} & +3.30 \\
& $90^{\circ}$ (Side) & 95.50 & \textbf{96.40} & +0.90 \\
& $180^{\circ}$ (Back) & 88.90 & \textbf{91.90} & +3.00 \\
\midrule
\multirow{2}{*}{Bag (BG)} 
& $0^{\circ}$--$180^{\circ}$ (All) & 91.37 & \textbf{91.75} & +0.38 \\
& $90^{\circ}$ (Side) & 92.00 & \textbf{93.40} & +1.40 \\
\midrule
\multirow{2}{*}{Coat (CL)} 
& $0^{\circ}$--$180^{\circ}$ (All) & 81.08 & \textbf{82.42} & +1.34 \\
& $90^{\circ}$ (Side) & 81.50 & \textbf{84.20} & +2.70 \\
\bottomrule
\end{tabular}
\end{table}

Gait recognition is highly sensitive to clothing and viewpoint variations. We show that joint actuation forces, as internal biomechanical signals, are more invariant to these external factors than appearance-based features. 

Using the CASIA-B dataset, we evaluate performance under clothing change (CL), carrying condition (BG), and viewpoint variation (Table~\ref{tab:casia_analysis_detailed}). The force modality yields the largest improvement under clothing change, where silhouettes are heavily distorted by coats (+1.34\%, nearly four times the BG gain of +0.38\%). Force also provides strong benefits at challenging non-frontal views, improving accuracy by +2.70\% under the $90^{\circ}$ CL condition, where self-occlusion degrades silhouette quality. Even under normal conditions, it helps disambiguate similar front/back views (+3.30\% at $0^{\circ}$, +3.00\% at $180^{\circ}$).


\subsubsection{Action Recognition: View and Subject Invariance}
\label{sec:view_vs_subject}
For action recognition, the goal is to learn representations invariant to viewpoint and subject identity while remaining sensitive to action class. We analyze the NTU-RGB+D datasets to examine how force contributes to these factors.

\begin{table}[t!]
\centering
\caption{\textbf{Action Recognition Robustness Evaluation on Cross-view NTU-60 Dataset.} We test generalization to each camera view after training on the other two. The official X-View protocol (testing on Cam 1) is shown, along with performance on the other two views. Accuracy (\%) is reported. Force consistently improves robustness to view changes.}
\label{tab:view_analysis_ntu60}
\resizebox{\columnwidth}{!}{%
\begin{tabular}{@{}llccc@{}}
\toprule
\textbf{Protocol} & \textbf{Training Views} & \textbf{Skeleton-Only} & \textbf{+ Force} & \textbf{$\Delta$} \\
\midrule
Test on Cam 1 (Official) & Cam 2, Cam 3 & 94.36 & \textbf{94.90} & +0.54 \\
Test on Cam 2 & Cam 1, Cam 3 & 93.05 & \textbf{93.88} & +0.83 \\
Test on Cam 3 & Cam 1, Cam 2 & 93.35 & \textbf{93.96} & +0.61 \\
\bottomrule
\end{tabular}}
\end{table}

\begin{table}[t!]
\centering
\caption{\textbf{Cross-subject And Cross-setup Analysis.} This table shows the performance on protocols designed to test generalization to new subjects and camera setups. Accuracy (\%) is reported. Force helps subject generalization but hurts setup generalization.}
\label{tab:subject_analysis_ntu}
\resizebox{\columnwidth}{!}{%
\begin{tabular}{@{}llccc@{}}
\toprule
\textbf{Invariance Type} & \textbf{Protocol (Dataset)} & \textbf{Skeleton-Only} & \textbf{+ Force} & \textbf{$\Delta$} \\
\midrule
\multirow{2}{*}{\textit{Subject}}
& X-Sub (NTU-60) & 89.53 & \textbf{89.96} & +0.43 \\
& X-Sub (NTU-120) & 85.55 & \textbf{85.86} & +0.31 \\
\midrule
\textit{Setup}
& X-Set (NTU-120)$^\dagger$ & \textbf{85.12} & 84.31 & -0.81 \\
\bottomrule
\multicolumn{5}{l}{$^\dagger$\footnotesize{X-Set evaluates generalization across camera and environment configurations}}
\end{tabular}}
\end{table}

\noindent\textbf{Force Enhances View Invariance.} 
As shown in Table~\ref{tab:view_analysis_ntu60}, adding force improves robustness to viewpoint changes. In the NTU-60 X-View protocol (test on Cam 1), accuracy increases by +0.54\% with similar gains (+0.83\% and +0.61\%) across the other two views. 

\noindent\textbf{Force Encodes Subject-specific Biometrics.}
Table~\ref{tab:subject_analysis_ntu} shows that force yields modest gains in cross-subject protocols (NTU-60 X-Sub: +0.43\%, NTU-120 X-Sub: +0.31\%) but a drop in NTU-120 X-Set (-0.81\%). This pattern aligns with gait recognition results in section \ref{sec:appearance_invariance}: force captures both action dynamics and individual execution style. When both subject and camera domains shift, the subject-specific component dominates, reducing generalization.

\noindent\textbf{Benefits For Dynamic Actions}
Force provides the greatest improvements for high-exertion actions, as shown in Table~\ref{tab:action_breakdown}. Accuracy for "punching/slapping" increases by +6.96\% and "throw" by +3.48\%, where rapid acceleration and peak torques offer strong discriminative cues. In-contrast, low-exertion actions like "reading" and "typing" show small declines, since limited joint torques contribute little meaningful signal. 

\begin{table}[t!]
\centering
\caption{\textbf{Per-action performance on NTU-60 X-View.} We report Top-1 accuracy (\%) for representative actions. Force provides the largest gains for high-exertion actions where dynamics are discriminative, while offering little benefit for subtle movements.}
\label{tab:action_breakdown}
\resizebox{\columnwidth}{!}{%
\begin{tabular}{@{}lccc@{}}
\toprule
\textbf{Action Class} & \textbf{Skeleton} & \textbf{+ Force} & \textbf{$\Delta$} \\
\midrule
\multicolumn{4}{c}{\textit{High-Exertion / Dynamic Actions}} \\
\midrule
A7: Throw & 91.46 & \textbf{94.94} & +3.48 \\
A24: Kicking something & 97.15 & \textbf{99.37} & +2.22 \\
A27: Jump up & 98.42 & \textbf{99.37} & +0.95 \\
A50: Punching/slapping & 81.65 & \textbf{88.61} & \textbf{+6.96} \\
\midrule
\multicolumn{4}{c}{\textit{Cyclical / Rhythmic Actions}} \\
\midrule
A26: Hopping & 96.52 & \textbf{97.47} & +0.95 \\
\midrule
\multicolumn{4}{c}{\textit{Low-Exertion / Static Actions}} \\
\midrule
A11: Reading & \textbf{83.81} & 79.68 & -4.13 \\
A12: Writing & 65.71 & \textbf{69.01} & +3.30 \\
A30: Typing on a keyboard & \textbf{81.65} & 75.95 & -5.70 \\
A33: Check time (from watch) & \textbf{96.52} & 97.15 & +0.63 \\
\midrule
\textbf{Avg. High-Exertion $\Delta$} & \multicolumn{3}{r}{\textbf{+3.40}} \\
\textbf{Avg. Low-Exertion $\Delta$} & \multicolumn{3}{r}{\textbf{-1.48}} \\
\bottomrule
\end{tabular}}
\end{table}

\subsubsection{Video Captioning: Reduced Hallucination with Physics Grounding}
\label{sec:captioning_qualitative}

\begin{figure*}[t]
    \centering
    \includegraphics[width=0.9\textwidth]{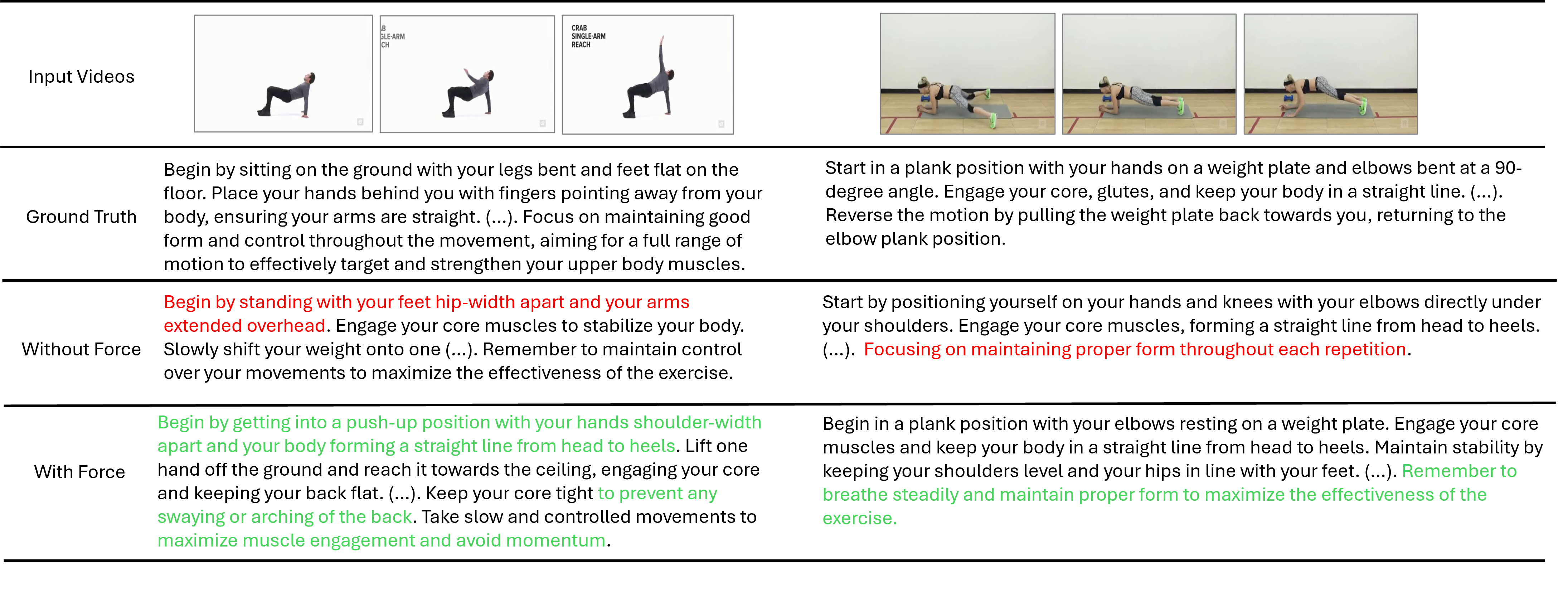}
    \caption{Comparison of Qwen2.5-VL on video captioning task. We compare captions generated by the baseline model (Without Force) and our force-augmented model (With Force). The green-highlighted words indicate phrases that explicitly capture biomechanical effort, while the red highlights mark imprecise phrases produced by the baseline model.}
    \label{fig:fig2}
\end{figure*}

Our quantitative results in Table~\ref{tab:overall_performance} showed that adding the force modality improved ROUGE-L (+0.029) and BERTScore (+0.0037). To understand \textit{how} force dynamics contribute to richer language, we present a qualitative comparison in Figure~\ref{fig:fig2}.
Across samples, we observed that the baseline model (Without Force) accurately describes the visible sequences of actions, however the force-agmented (With Force) additionally captures the physical intent underlying those movements. For instances, in the "Crap Single-Arm Reach" sample, the baseline provide a correct description (e.g "lift your hips", "extend one arm"), while the force-augmented model generates captions that are both richer and more physically aware, such as "engaging your core", "prevent any swaying or aching of the back", and "maximize muscle movement". Similarly, in the "Plank Jack" example, the baseline correctly identifies the "straight plank position." Our "With Force" model adds the crucial biomechanical instruction to maintain this pose by "breath steadily and maintain proper form" which implies the active muscular stabilization (i.e., force exertion) required to perform the movement without compromising the plank. These examples demonstrate that the force modality provides a new, physics-based context that allows the VLM to generate captions that are more physically grounded.

\subsection{Ablation Studies}
\label{sec:ablation}

Finally, we systematically ablate design choices to identify optimal fusion strategies and validate key components.

\subsubsection{Ablation on Fusion Strategy}
\label{sec:fusion_ablation}
The fusion mechanism, introduced in Section~\ref{sec:gait_task}, combines silhouette appearance features ($\phi_{sil}$) and force embeddings ($\phi_{\text{force}}$) at the feature level to align spatial and dynamic cues. Table~\ref{tab:fusion_ablation_gait} summarizes results across three gait benchmarks. Simple fusion methods such as addition, concatenation, or element-wise multiplication provide minimal or negative gains, showing that naive operations cause feature interference. Learnable approaches like Gated Attention yield moderate improvement, while the BMM (Spatial Transform) operator consistently achieves the best results (47.3\% on Gait3D, 20.6\% on CCGR-mini, 90.39\% on CASIA-B). This indicates that adaptive, spatially aware fusion most effectively aligns force dynamics with visual features and generalizes across datasets.

\begin{table}[t!]
\centering
\caption{\textbf{Fusion Strategy Ablation.} We report Rank-1 Accuracy (\%) for Gait3D and CCGR-mini, and Rank-1 (Avg.) \% for CASIA-B. }
\label{tab:fusion_ablation_gait}
\resizebox{\columnwidth}{!}{%
\setlength{\tabcolsep}{4pt} 
\begin{tabular}{@{}lccc@{}}
\toprule
\textbf{Fusion Strategy (Sil + Force)} & \textbf{Gait3D} & \textbf{CCGR-mini} & \textbf{CASIA-B} \\
\midrule
Addition & 45.4 & 19.8 & 89.81 \\
Concatenation & 44.8 & 19.5 & 89.72 \\
Element-wise Multiplication & 42.8 & 19.1 & 89.15 \\
Gated Attention & 46.3 & 20.2 & 90.17 \\
\textbf{BMM (Spatial Transform)} & \textbf{47.3} & \textbf{20.6} & \textbf{90.39} \\
\bottomrule
\end{tabular}%
} 
\end{table}

\subsubsection{Ablation on Model Architecture}
\label{sec:model_variant_ablation}
To evaluate whether the benefits of the force modality generalize beyond specific backbones, we replace the primary architectures in gait and action recognition with alternative models and report their performance.

\begin{table}[t]
\centering
\caption{\textbf{Ablation On Different Model Architectures Across Tasks And Datasets.} 
Adding the force modality improves performance on most backbones, 
highlighting its generalization ability.}
\label{tab:model_variants}
\resizebox{\columnwidth}{!}{%
\begin{tabular}{@{}llllcc@{}}
\toprule
\textbf{Task} & \textbf{Dataset} & \textbf{Backbone} & \textbf{Baseline} & \textbf{+ Force} & \textbf{$\Delta$} \\
\midrule
\multirow{4}{*}{Gait} 
& CASIA-B & SMPLGait~\cite{zheng2022gait} & 89.52 & 90.39 & +0.87 \\
& CASIA-B & GaitBase~\cite{chao2019gaitset} & 89.6 & 89.82 & +0.22 \\
& Gait3D & SMPLGait~\cite{zheng2022gait} & 46 & 47.3 & +1.30 \\
& Gait3D & SkeletonGait++~\cite{fan2024skeletongait} & 74.90 & 73.70 & -1.2 \\
\midrule
\multirow{2}{*}{Action} 
& NTU-60 XV & CTR-GCN~\cite{chen2021channel} & 94.36 & 94.90 & +0.54 \\
& NTU-60 XV & BlockGCN~\cite{zhou2024blockgcn} & 92.69 & 93.11 & +0.42 \\
\midrule
\multirow{3}{*}{Captioning} 
& BoFiT & InternVL-3~\cite{zhu2025internvl3} & 0.415 & 0.429 & +0.014 \\
& BoFiT & Qwen2.5-VL~\cite{bai2025qwen2} & 0.310 & 0.339 & +0.029 \\
\bottomrule
\end{tabular}}
\end{table}

As shown in Table~\ref{tab:model_variants}, the force modality consistently improves performance across most architectures: GaitBase (+0.22\% on CASIA-B), BlockGCN (+0.42\% on NTU-60), and InternVL3 for captioning. However, for SkeletonGait++, where adding force causes a drop (–1.2\% on Gait3D). We attribute this to a modality interaction mismatch—SkeletonGait++ already integrates silhouettes and 2D joint heatmaps, which partially encode dynamic cues. In this case, force signals may overlap with existing motion information, introducing redundancy.

\subsubsection{Ablation on Individual Modalities}
\label{sec:component_ablation}

\begin{table}[t!]
\centering
\caption{\textbf{Multi-modal Component Analysis On Gait3D.} Each modality contributes complementary information. }
\label{tab:component_ablation}
\small
\begin{tabular}{@{}cccccc@{}}
\toprule
\textbf{Silhouette} & \textbf{Force} & \textbf{SMPL} & \textbf{Rank-1} & \textbf{mAP} & \textbf{mINP} \\
\midrule
\checkmark &            &            & 46.0 & 38.02 & 23.08 \\
\checkmark & \checkmark &            & 47.3 & 39.11 & 24.00 \\
\checkmark &            & \checkmark & 49.2 & 40.68 & 25.55 \\
\checkmark & \checkmark & \checkmark & \textbf{51.9} & \textbf{41.59} & \textbf{24.79} \\
\bottomrule
\end{tabular}
\end{table}

Table~\ref{tab:component_ablation} shows ablations of individual modalities on Gait3D. Critically, three-stream fusion (Sil + Force + SMPL) outperforms other combinations, confirming that force provides complementary information on the Gait3D dataset.

\section{Conclusion}

In summary, we investigated when and how physics-inferred forces improve human motion understanding across recognition and captioning tasks. Across eight benchmarks (Gait3D, CASIA-B, CCGR-mini, NTU-60/120, NW-UCLA, PennAction, and BoFiT), force features consistently complement appearance and kinematics. Our analyses show that force improves robustness to appearance changes and cross-view shifts. Fusion strategies are task dependent, where spatial transforms benefit dense gait features, while late or embedding-level fusion works better for graph and language models. Dynamic actions gain the most from force cues, and language models enriched with force generate more semantically and physically grounded descriptions. These findings establish force dynamics as a general, complementary modality that enhances robustness across diverse tasks.


{
    \small
    \bibliographystyle{ieeenat_fullname}
    \bibliography{main}
}

\appendix
\clearpage
\setcounter{page}{1}
\maketitlesupplementary

\appendix

\noindent\textbf{Appendix} 
\begin{itemize}
    \item Section \ref{app:settings}: Experimental Settings
    \item Section \ref{sec:stat_significance}: Statistical Significance Analysis
    \item Section \ref{app:joint}:  Detailed Analysis on Joint Influence
    \item Section \ref{app:caption results}:  Additional Video Captioning Samples
    \item Section \ref{app:ethics}: Ethical Considerations
\end{itemize}

\section{Experimental Settings}
\label{app:settings}
\subsection{Implementation Details}
\label{app:impl}
We provide implementation details to facilitate reproducibility. 
\begin{itemize}
  \item \textbf{Force Network (FN)} 
  The FN consists of a three-layer MLP (hidden dimensions: 128–256–256) followed by LayerNorm and GELU activation. 
  The resulting feature $\phi_{\text{force}}$ is projected into the shared embedding space via a linear layer.
  \item \textbf{Training Schedules and Hardware.}
  All experiments were conducted on 4×NVIDIA L40S GPUs with mixed precision.  
\end{itemize}

\subsection{Dataset Processing}
\label{app:data}
\begin{itemize}
  \item \textbf{NTU RGB+D} 
  datasets are represented using a unified skeleton format with 25 keypoints per person, following the standard NTU RGB+D topology. We utilize a dual-stream input consisting of bone vectors (joint displacement pairs) and physics-inferred force vectors. Each clip is temporally resampled to 64 frames and normalized within a fixed spatial range. For robustness, training applies random rotation, temporal cropping, and spatial jittering. During testing, sequences are uniformly sampled without augmentation.
  \item \textbf{BoFiT.} 
  We employ a subject-independent split stratified by exercise level, ensuring balanced label distributions across training and testing sets, resulting in 2000 training and 360 testing videos.
  \item \textbf{Force Signals.} 
  Raw force magnitudes were normalized per sequences before feeding into the model.
\end{itemize}

\subsection{Fusion Operator Designs}
\label{app:fusion}
This subsection specifies the mathematical definitions and implementation details of the fusion operators used in our gait recognition ablation study (Table 6 in the main paper). While Table 6 reports the quantitative performance of each operator, here we clarify how each fusion function ${F}$ acts on the visual and force features, ensuring that our experimental setup is precise and reproducible. Beyond simply stating which operator yields the best accuracy, we systematically compare different designs of ${F}$ between visual and force features to understand how the choice of fusion mechanism affects performance.

\begin{itemize}
  \item \textbf{Addition:} $F_{\text{fused}} = F_{\text{sil}} + \text{Proj}(\phi_{\text{force}})$.
  \item \textbf{Concatenation:} $F_{\text{fused}} = \text{Conv}_{1\times1}([F_{\text{sil}}; \text{Proj}(\phi_{\text{force}})])$, where a $1\times1$ convolution learns to merge concatenated features.
  \item \textbf{Element-wise Multiplication:} $F_{\text{fused}} = F_{\text{sil}} \odot \text{Proj}(\phi_{\text{force}})$, enabling force to modulate appearance activations dynamically.
  \item \textbf{Spatial Transformation (BMM):} 
  Following SMPLGait, $\phi_{\text{force}}$ generates an affine transformation $G_{\text{force}}\in\mathbb{R}^{w\times w}$ applied to $F_{\text{sil}}$ via batched matrix multiplication:
  $\hat{F}_{\text{sil}} = \text{BMM}(F_{\text{sil}}, I + G_{\text{force}})$.
  This operation spatially normalizes features using force cues.
  \item \textbf{Gated Attention:} 
  An attention weight $\alpha = \sigma(\text{Linear}([\text{GAP}(F_{\text{sil}}); \phi_{\text{force}}]))$ adaptively balances each modality’s contribution.
\end{itemize}

\section{Statistical Significance Analysis}
\label{sec:stat_significance}

To validate that the performance gains reported in Section 6 arise from the systematic integration of force dynamics rather than stochastic training variance, we conduct a rigorous statistical significance analysis. For each task, we train both the baseline (without force) and our force-augmented model independently across 5 random seeds ($S \in \{42, 10, 100, 1000, 10000\}$) and evaluate them on identical test splits.

We employ an independent two-sample $t$-test to compare the baseline and our method. Let $\{\bar{X}_1, s_1^2, n_1\}$ and $\{\bar{X}_2, s_2^2, n_2\}$ denote the sample means, variances, and number of runs for the baseline and force-augmented models, respectively. The $t$-statistic is computed as:
\begin{equation}
    t = \frac{\bar{X}_1 - \bar{X}_2}{\sqrt{\frac{s_1^2}{n_1} + \frac{s_2^2}{n_2}}}
\end{equation}
where $n_1 = n_2 = 5$. We test the null hypothesis $H_0$ that both methods have equal expected performance against a significance level of $\alpha = 0.05$. Improvements yielding $p < 0.05$ are considered statistically significant.

\subsection{Gait Recognition (CASIA-B)}
\label{sec:stat_gait}

\noindent\textbf{Experimental Setup.}
We utilize the CASIA-B dataset with the SMPLGait framework. To ensure a fair comparison, we employ the same optimization protocol for both the silhouette-only baseline and the force-augmented model. Training is conducted for $100{,}000$ iterations using an SGD optimizer with a momentum of $0.9$ and weight decay of $5 \times 10^{-4}$. The learning rate is initialized at $0.2$ and follows a MultiStepLR schedule, decaying by a factor of $0.1$ at iterations $20\text{k}$, $40\text{k}$, and $50\text{k}$. We employ a balanced sampler with a batch size of $(8 \times 16)$—denoting 8 identities with 16 sequences each—and optimize using a combined Triplet Loss (margin $0.2$) and Cross-Entropy Loss.

\noindent\textbf{Significance Results.}
We report the average Rank-1 accuracy over the Normal (NM), Bag (BG), and Coat (CL) conditions. As detailed in Table~\ref{tab:stat_gait}, the baseline model achieves a mean accuracy of $89.55 \pm 0.05\%$. The integration of force dynamics raises this to $90.35 \pm 0.06\%$. The resulting $p$-value is $< 0.001$, indicating that the improvement is highly robust to initialization noise.

\begin{table}[h]
\centering
\caption{\textbf{Statistical Significance on Gait Recognition (CASIA-B).} We report Rank-1 Accuracy (\%) averaged over 5 runs. The improvement is statistically significant ($p < 0.001$).}
\label{tab:stat_gait}
\small
\begin{tabular}{@{}lccc@{}}
\toprule
\textbf{Metric} & \textbf{Baseline} & \textbf{+ Force} & \textbf{$p$-value} \\
\midrule
Rank-1 (Avg.) & $89.55 \pm 0.05$ & $\mathbf{90.35 \pm 0.06}$ & $< 0.001$ \\
\bottomrule
\end{tabular}
\end{table}

\subsection{Action Recognition (NTU-60)}
\label{sec:stat_action}

\noindent\textbf{Experimental Setup.}
Experiments are conducted on the NTU-RGB+D 60 dataset using the Cross-View (X-View) protocol. We employ the CTR-GCN backbone, trained for $60$ epochs with a batch size of $64$. Optimization is performed using SGD with Nesterov momentum ($0.9$) and weight decay of $3 \times 10^{-4}$. The initial learning rate is set to $0.1$ with a step decay strategy, reducing the rate by a factor of $10$ at epochs $40$ and $50$.

\noindent\textbf{Significance Results.}
We compare the Top-1 Classification Accuracy in Table~\ref{tab:stat_action}. The Skeleton-Only baseline yields $94.40 \pm 0.24\%$, while the force-augmented variant consistently outperforms it, reaching $94.97 \pm 0.09\%$. The calculated $p$-value of $0.001$ confirms that force dynamics provide a consistent discriminative signal for action classification.

\begin{table}[h]
\centering
\caption{\textbf{Statistical Significance on Action Recognition (NTU-60 X-View).} Top-1 Accuracy (\%) averaged over 5 runs. The improvement is statistically significant ($p = 0.001$).}
\label{tab:stat_action}
\small
\begin{tabular}{@{}lccc@{}}
\toprule
\textbf{Metric} & \textbf{Baseline} & \textbf{+ Force} & \textbf{$p$-value} \\
\midrule
Accuracy (\%) & $94.40 \pm 0.24$ & $\mathbf{94.97 \pm 0.09}$ & $0.001$ \\
\bottomrule
\end{tabular}
\end{table}

\subsection{Fine-Grained Video Captioning (BoFiT)}
\label{sec:stat_caption}

\noindent\textbf{Experimental Setup.}
We evaluate fine-grained video captioning on the BoFiT dataset using the Qwen2.5-VL-3B-Instruct architecture. To isolate the contribution of force cues, we freeze the vision encoder and finetune only the projector and LLM parameters. Training runs for $50$ epochs with a global batch size of $64$ (accumulated over 4 steps on 4 GPUs). We use the AdamW optimizer with a cosine learning rate scheduler, a peak learning rate of $2 \times 10^{-7}$, and a warmup ratio of $0.03$. All experiments utilize \texttt{bfloat16} precision.

\noindent\textbf{Significance Results.}
Table~\ref{tab:stat_caption} summarizes the results for ROUGE-L and BERTScore. The force-augmented model demonstrates statistically significant improvements across both metrics: ROUGE-L increases from $0.307$ to $0.339$ ($p < 0.001$), and BERTScore improves from $0.889$ to $0.896$ ($p = 0.002$). The low standard deviations confirm that force cues systematically enhance the semantic quality of generated captions.

\begin{table}[h]
\centering
\caption{\textbf{Statistical Significance on Video Captioning (BoFiT).} We report Mean $\pm$ Standard Deviation over 5 runs. Both metrics show statistically significant improvements.}
\label{tab:stat_caption}
\small
\begin{tabular}{@{}lccc@{}}
\toprule
\textbf{Metric} & \textbf{Baseline} & \textbf{+ Force} & \textbf{$p$-value} \\
\midrule
ROUGE-L & $0.307 \pm 0.004$ & $\mathbf{0.339 \pm 0.005}$ & $< 0.001$ \\
BERTScore & $0.889 \pm 0.003$ & $\mathbf{0.896 \pm 0.001}$ & $0.002$ \\
\bottomrule
\end{tabular}
\end{table}

\section{Detailed Analysis on Joint Influence}
\label{app:joint}

\begin{figure*}[t]
    \centering
    \includegraphics[width=0.8\textwidth]{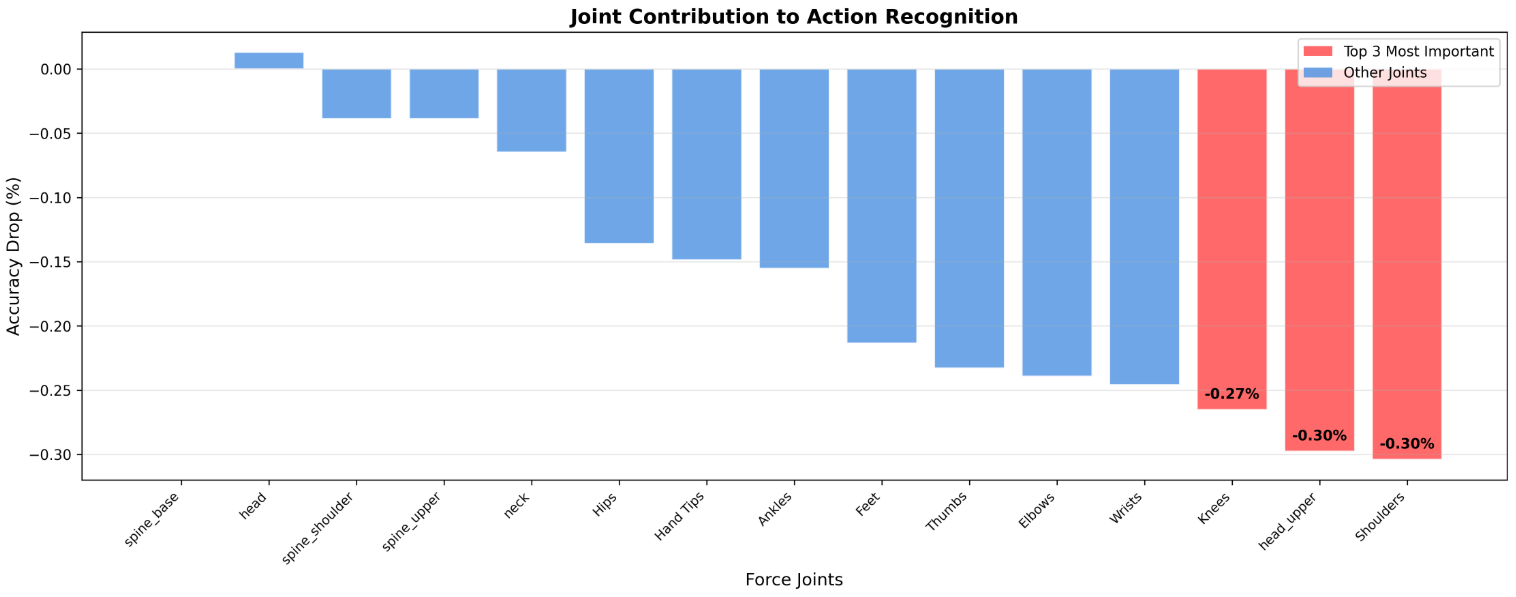}
    \caption{Joint-level force ablation on the NTU-RGB+D 60 dataset. Accuracy decreases are measured when masking each joint's force feature from the action recognition network. The model is most sensitive to forces around the shoulders, upper head, and knees, highlighting that upper-body and supporting-limb dynamics play key roles in distinguishing actions.}
    \label{fig:joint1}
\end{figure*}

\begin{figure*}[t]
    \centering
    \includegraphics[width=0.8\textwidth]{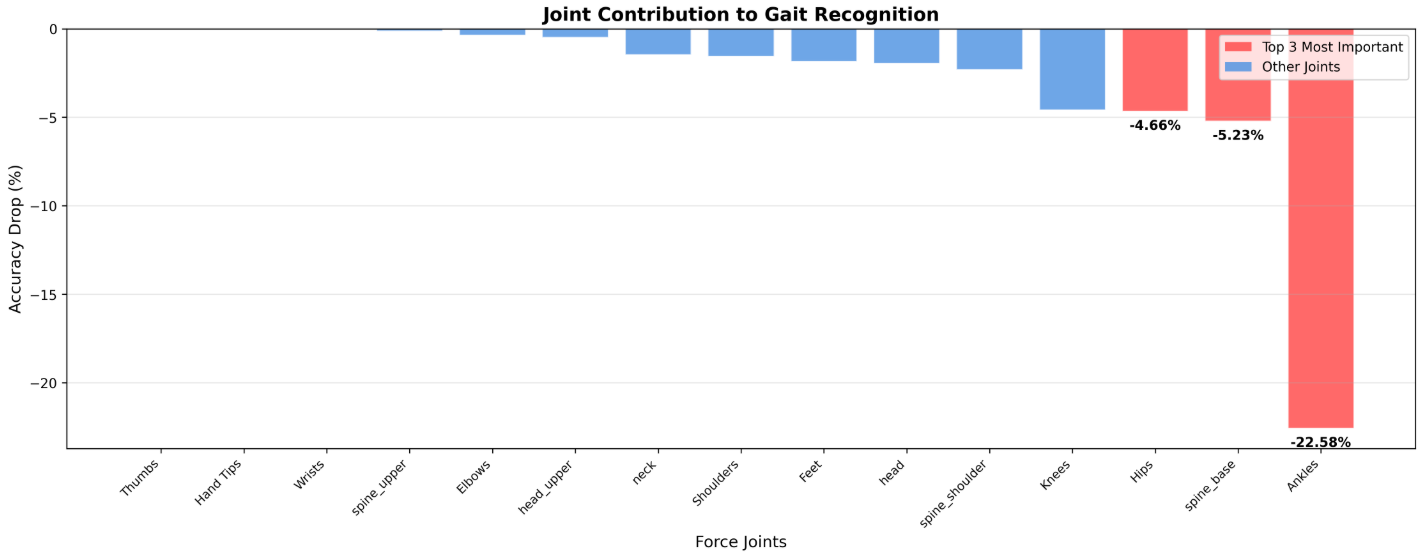}
    \caption{Joint-level force ablation on the CASIA-B gait dataset. Accuracy decreases are measured when masking each joint's force feature from gait-recognition network. The greatest degradation occurs at the ankles, spine base and hips, showing that lower-body propulsion and balance forces dominate gait identity cues, while upper-body joints contribute less.}
    \label{fig:joint2}
\end{figure*}

\begin{figure*}[ht!]
    \centering
    \includegraphics[width=0.8\textwidth]{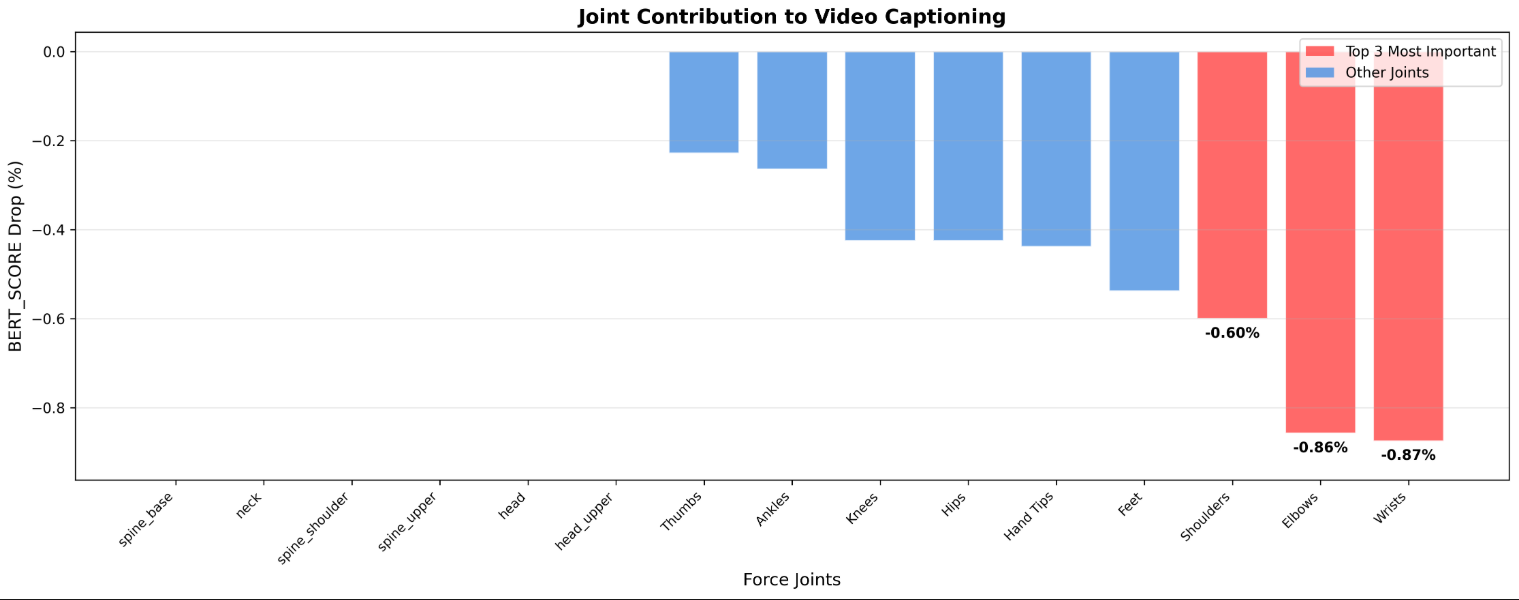}
    \caption{Joint-level force ablation on the BoFiT video captioning dataset. Each bar represents the drop in BERTScore when the force feature of the corresponding joint is masked. Larger negative drops indicate greater importance to caption quality. The most influential joints are shoulders, elbows, and wrists, highlighting that upper-body forces provide key cues for describing fine-grained human motion in natural-language video understanding.}
    \label{fig:joint3}
\end{figure*}

We further analyze the contribution of individual torque joints by masking their corresponding force features and observing the resulting performance degradation, as illustrated in Figures~\ref{fig:joint1}, \ref{fig:joint2}, and \ref{fig:joint3}. These joint-level ablations reveal distinct dependency patterns across tasks:
\begin{itemize}
    \item \textbf{Action Recognition (NTU RGB+D 60):} The largest accuracy drops occur when masking the shoulders, upper head, and knees (Figure \ref{fig:joint1}), indicating that upper-body forces and supporting-limb dynamics are crucial for distinguishing human actions involving arm swings, body tilts, or locomotion. This aligns with how torque from these regions captures the intent and directionality of motion beyond skeletal pose.

    \item \textbf{Gait Recognition (CASIA-B):} As shown in Figure \ref{fig:joint2}, performance degrades most at the ankles, spine base, and hips, emphasizing that lower-body propulsion and balance forces dominate gait identity cues. In contrast, upper-body forces (e.g., shoulders, head) have limited influence, suggesting that gait recognition primarily relies on the periodic stability of lower-limb torques.

    \item \textbf{Video Captioning (BoFiT):} In Figure \ref{fig:joint3}, the most influential joints are the shoulders, elbows, and wrists, reflecting that upper-body forces contribute to describing fine-grained hand and arm movements in natural-language grounding. These force cues enrich the semantic content of captions by signaling exertion, interaction, and motion emphasis.
\end{itemize}

\begin{figure*}[ht!]
    \centering
    \includegraphics[width=1.0\textwidth]{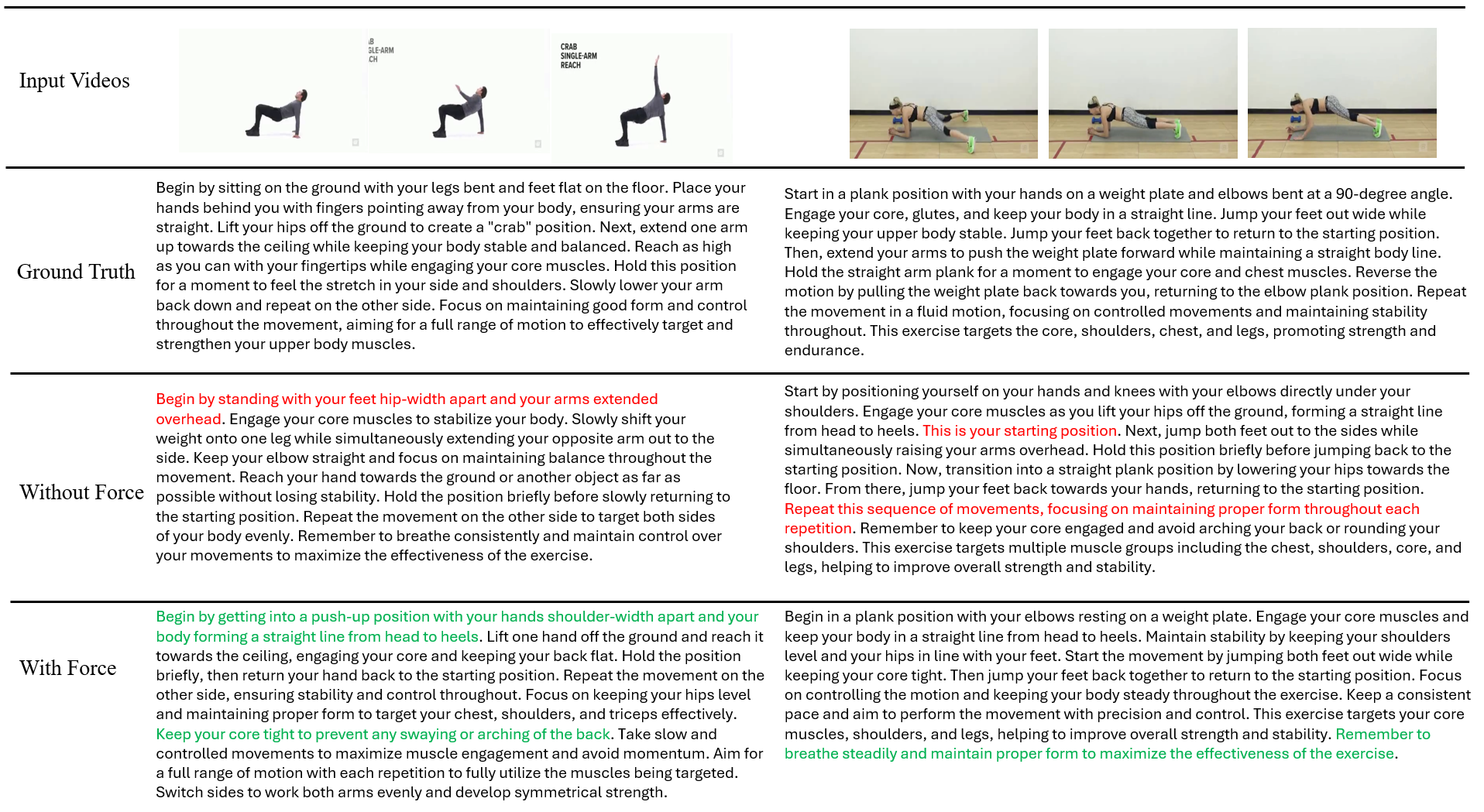}
    \caption{Captioning examples from the BoFiT dataset. Left: Crab single-arm reach exercise. Right: Lying leg curls exercise.}
    \label{fig:cap1}
\end{figure*}

\begin{figure*}[ht!]
    \centering
    \includegraphics[width=1.0\textwidth]{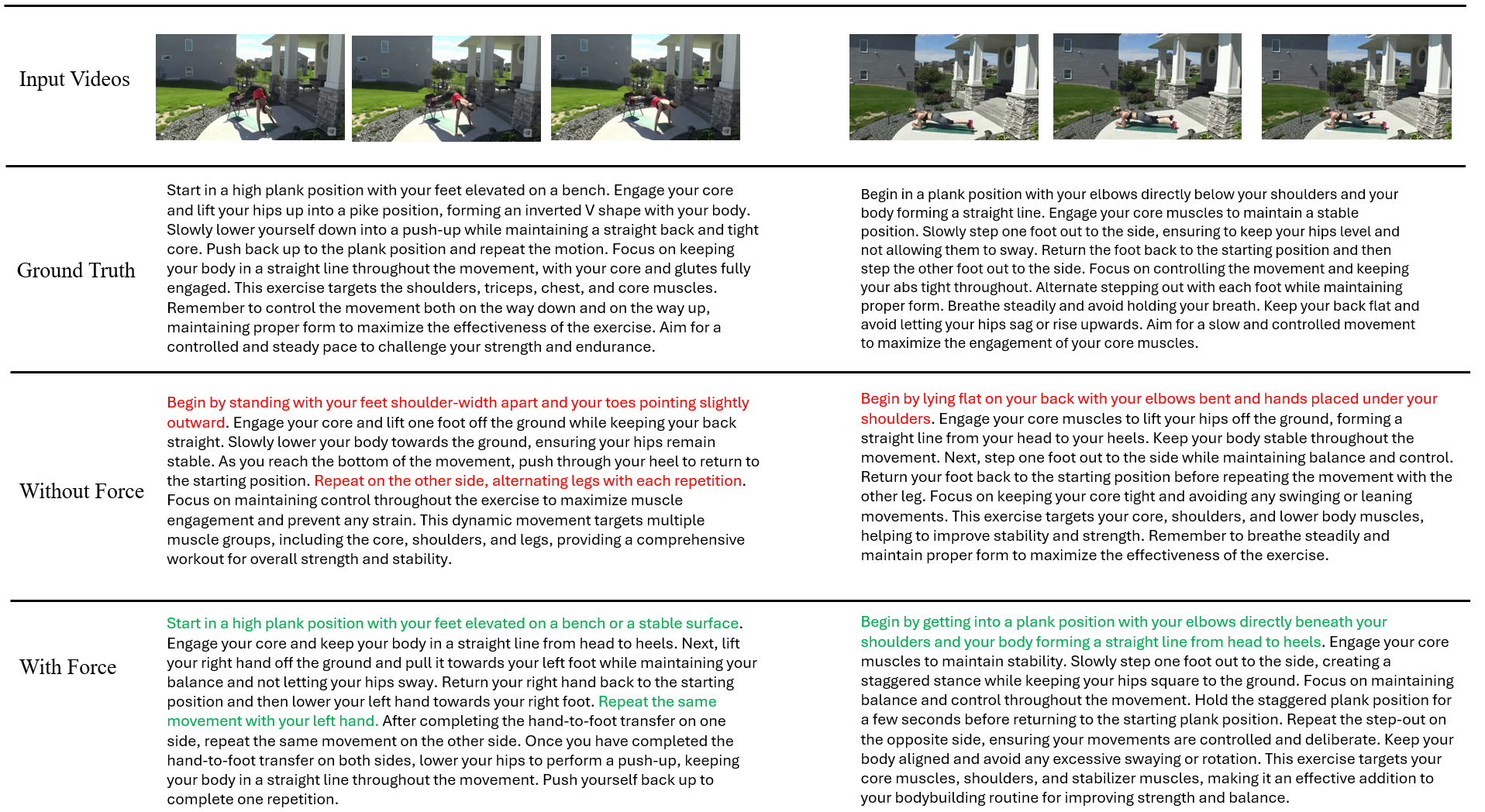}
    \caption{Captioning examples from the BoFiT dataset. Left: Elevated plank–pike–push-up exercise. Right: Elbow plank with step-out movement.}
    \label{fig:cap2}
\end{figure*}

\section{Additional Video Captioning Samples}
\label{app:caption results}

Figure~\ref{fig:cap1} in the supplementary illustrates detailed captioning outputs corresponding to Figure 3 in the main paper. Additional qualitative examples of our video captioning model are shown in Figure~\ref{fig:cap2}.

\section{Ethical Considerations}
\label{app:ethics}
This work relies on publicly available datasets collected under prior consent agreements.
While our models focus on understanding movement physics rather than identifying individuals, 
force-augmented gait recognition may raise privacy concerns. 
We advocate for dataset anonymization, synthetic data augmentation, and controlled-access evaluation to mitigate misuse in surveillance contexts.

\end{document}